\documentclass[10pt,twocolumn,letterpaper]{article}

\usepackage{cvpr}      % To produce the REVIEW version

\usepackage{multirow}
\usepackage{colortbl}
\usepackage{color}
\usepackage[table]{xcolor} % 引入xcolor包，并使用table选项以支持colortbl功能
\usepackage{gradient-text}
\usepackage{xspace}
\usepackage{marvosym}
\newcommand\NickName{{\gradientRGB{SpatialSky}{255, 106, 0}{192, 0, 0}}\xspace}

\definecolor{myblue}{rgb}{0.88,0.98,1}
\definecolor{mygreen}{rgb}{0.92, 1.0, 0.92}
\definecolor{myred}{rgb}{1, 0.9, 0.9}

\definecolor{cvprblue}{rgb}{0.21,0.49,0.74}
\usepackage[pagebackref,breaklinks,colorlinks,allcolors=cvprblue]{hyperref}

\def\paperID{1825} % *** Enter the Paper ID here
\def\confName{CVPR}
\def\confYear{2026}

\title{\raisebox{-2mm}{\includegraphics[width=1cm]{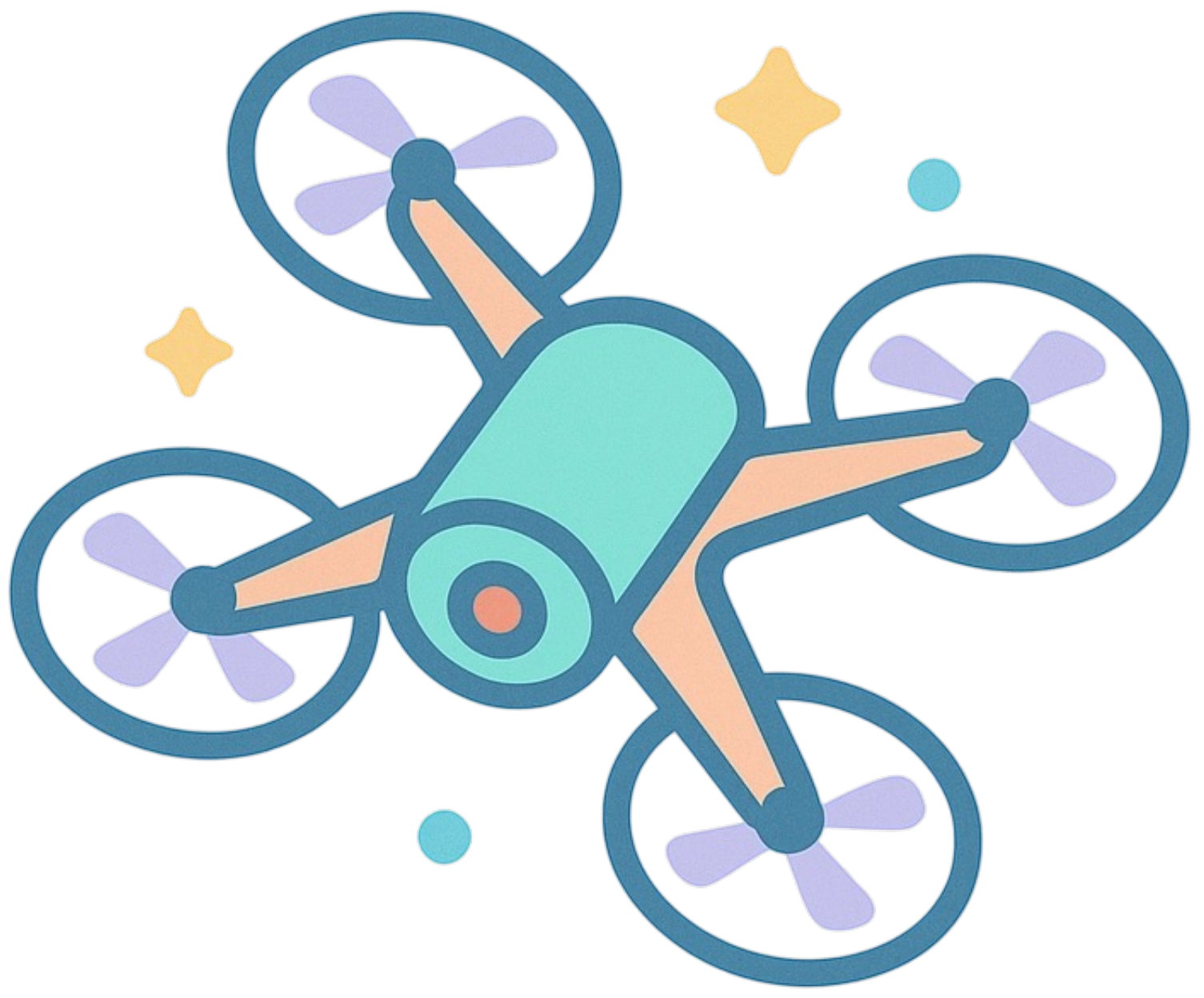}}~~Is your VLM Sky-Ready? A Comprehensive Spatial Intelligence \\ Benchmark for UAV Navigation}

\author{\bf  Lingfeng Zhang$^{1,2,3,*}$, Yuchen Zhang$^{3,4,*}$,  Hongsheng Li$^{1}$, Haoxiang Fu$^{6}$, Yingbo Tang$^{5}$\\
\bf Hangjun Ye$^{3}$, Long Chen$^3$, Xiaojun Liang$^2$, Xiaoshuai Hao$^{3,\dagger,\text{\Letter}}$, Wenbo Ding$^{1, \text{\Letter}}$ 
\vspace{0.5em}
\\
$^1$  Tsinghua Shenzhen International Graduate School, Tsinghua University \\ 
$^2$ Peng Cheng Laboratory 
$^3$  Xiaomi EV 
$^4$  Georgia Institute of Technology \\
$^5$  Institute of Automation, CAS 
$^6$ National University of Singapore\\
\tt \small zlf25@mails.tsinghua.edu.cn, haoxiaoshuai@xiaomi.com, ding.wenbo@sz.tsinghua.edu.cn
}

\begin{document}
% \maketitle
\twocolumn[{
\renewcommand\twocolumn[1][]{#1}
% \vspace{-50pt}
\maketitle
% \vspace{-2em}

\vspace{-30pt}
\begin{center}
    \captionsetup{type=figure}
    % \vspace{-32pt}
    \includegraphics[width=0.98\textwidth]{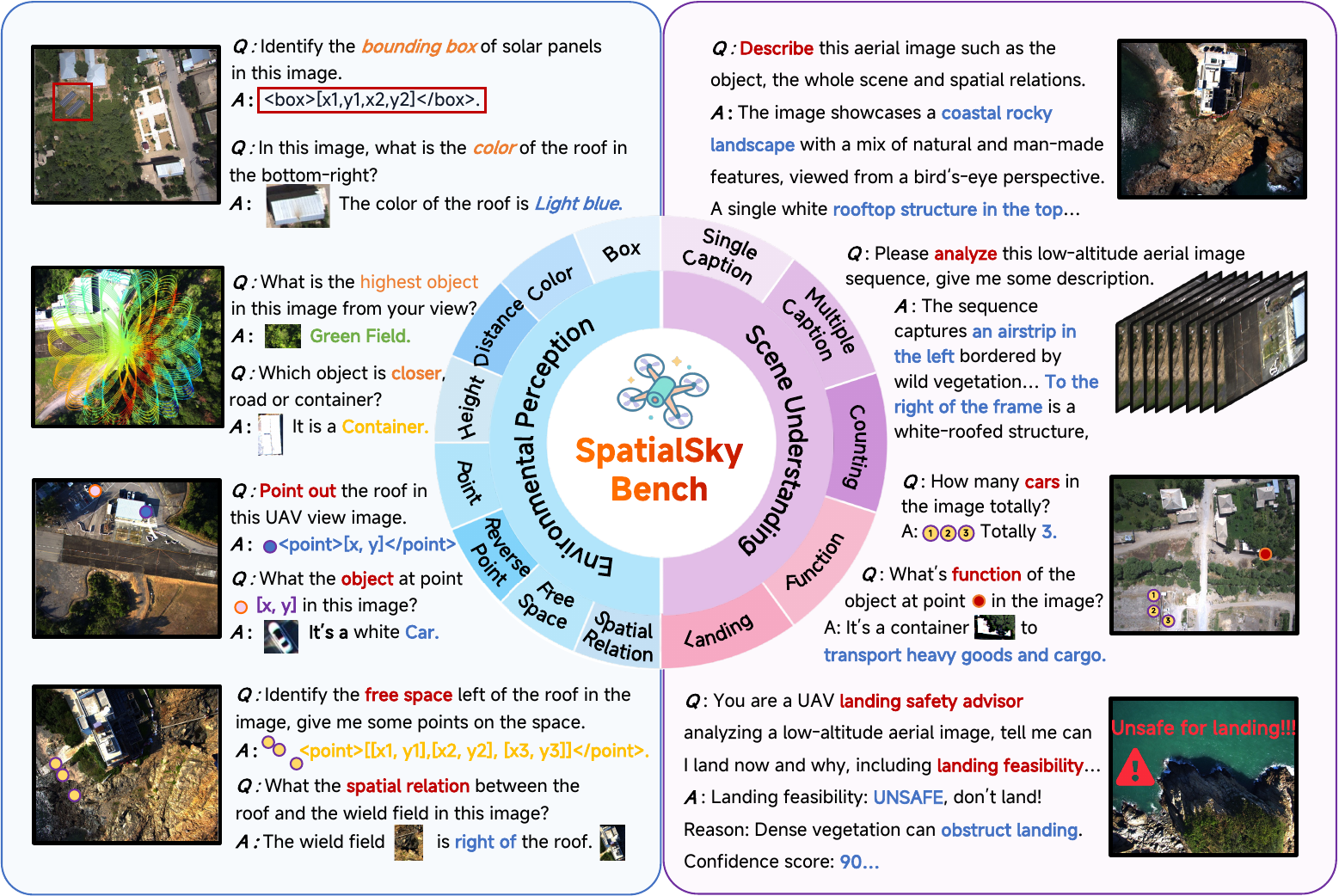}
    % \vspace{-10pt}
\captionof{figure}{\textbf{Overview of \NickName-Bench.} 
Our benchmarks are divided into two categories: Environmental Perception and Scene Understanding, covering a total of 13 subcategories. We evaluated the VLM’s spatial intelligence capabilities across these UAV navigation tasks.
}
    \label{fig1}
\end{center}
% \vspace{-3pt}
}]

\let\thefootnote\relax\footnotetext{$^{*}$ Equal contribution.}
\let\thefootnote\relax\footnotetext{$^{\dagger}$ Project leader.}
\let\thefootnote\relax\footnotetext{$^{\text{\Letter}}$ Corresponding authors.}

\begin{abstract}
Vision-Language Models (VLMs), leveraging their powerful visual perception and reasoning capabilities, have been widely applied in Unmanned Aerial Vehicle (UAV) tasks.
%%%%%%
However, the spatial intelligence capabilities of existing VLMs in UAV scenarios remain largely unexplored, raising concerns about their effectiveness in navigating and interpreting dynamic environments.
%%%%%%
To bridge this gap, we introduce \textbf{\NickName-Bench}, a comprehensive benchmark specifically designed to evaluate the spatial intelligence capabilities of VLMs in UAV navigation. 
%%%%%%
Our benchmark comprises two categories—Environmental Perception and Scene Understanding—divided into 13 subcategories, including bounding boxes, color, distance, height, and landing safety analysis, among others.
%%%%%%
Extensive evaluations of various mainstream open-source and closed-source VLMs reveal unsatisfactory performance in complex UAV navigation scenarios, highlighting significant gaps in their spatial capabilities.
%%%%%%
To address this challenge, we developed the \textbf{\NickName-Dataset}, a comprehensive dataset containing 1 M samples with diverse annotations across various scenarios. 
Leveraging this dataset, we introduce \textbf{\textit{Sky-VLM}}, a specialized VLM designed for UAV spatial reasoning across multiple granularities and contexts.
%%%%%%
Extensive experimental results demonstrate that \textbf{Sky-VLM}  achieves state-of-the-art performance across all benchmark tasks, paving the way for the development of VLMs suitable for UAV scenarios.
%%%%%%
The source code is available at~\url{https://github.com/linglingxiansen/SpatialSKy} .
% The dataset, benchmark toolkit, and associated code and model checkpoints will be publicly accessible.

\end{abstract}    
\section{Introduction}
\label{sec:intro}

Recently, the rapid development of Vision-Language Models (VLMs) has demonstrated their remarkable ability to understand and reason about visual scenes~\cite{tang2025roboafford,wuevaluating,Robospatial,ji2025robobrain,fu2024blink,tanreason,tang2025affordgrasp,hao2022listen,hao2021matters,zhang2025vtla,wu2025evaluating,team2025robobrain,zhang2025video,cheng2025exploring,zhang2025humanoidpano}. With the increasing prevalence of unmanned aerial vehicles (UAVs) in search and rescue operations, infrastructure inspection, and precision agriculture, VLMs have been successfully applied to UAV visual navigation tasks~\cite{aerialvln,openuav,uavvla,seepointfly,soranav,yaqoot2025uavvlrrvisionlanguageinformednmpc,li2025skyvln,zhang2024trihelper,zhang2025multi,zhang2025nava,liu2025toponav,zhang2025team}, showing promising application prospects.
% Spatial intelligence capabilities of VLMs are curial for UAV navigation, which provide: detailed understanding of spatial relationships, fine-grained scene understanding, and precise environmental perception. These spatial reasoning capabilities can support real-time autonomous navigation decisions.
The spatial intelligence of VLMs is crucial for UAV navigation, enabling a detailed understanding of spatial relationships, fine-grained scene understanding, and precise environmental perception to support real-time UAV navigation decisions.
However, existing VLM evaluation benchmarks primarily focus on human perspectives, such as indoor scenes, street scenes, and images taken with handheld cameras~\cite{VSI, mmsi, dang2025rynnec, zhou2025roborefer,wu2025evaluating,zhang2025video,li2025vquala,zheng2025railway,gong2025stairway,xiao2025team,zheng2025generation}. This difference in perspective makes existing benchmarks unable to assess the spatial intelligence capabilities of VLMs in UAV scenarios.

\begin{figure}
\centering
    \captionsetup{type=figure}
    % \vspace{-32pt}
    \includegraphics[width=0.48\textwidth]{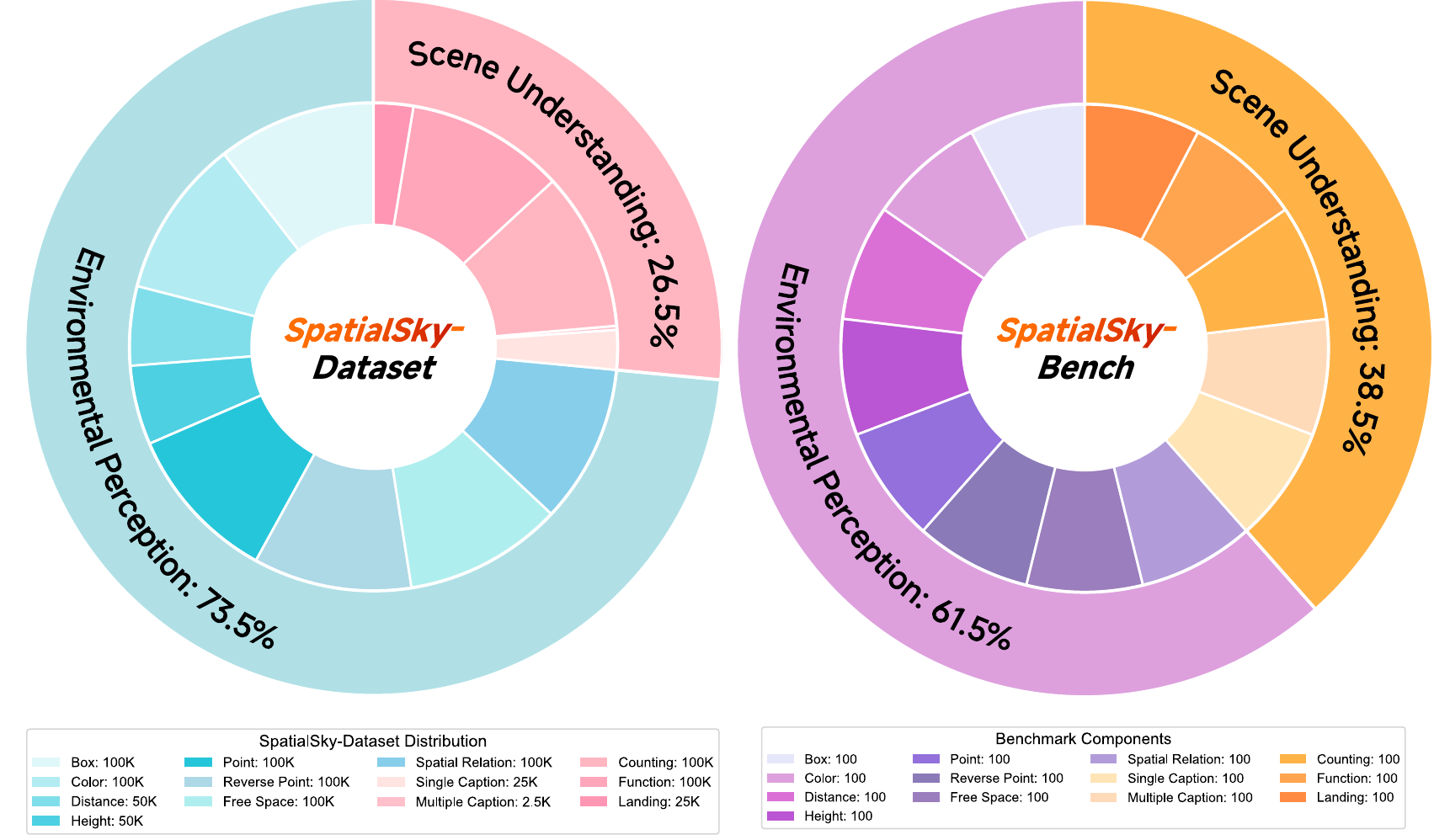}
    % \vspace{-10pt}
% \captionof{figure}{\textbf{\textit{\NickName-Dataset} and \textit{\NickName-Bench} Distribution.} }
\captionof{figure}{\textbf{Distribution of our dataset and benchmark.} }
\vspace{-1em}
    \label{fig: distribution}
\end{figure}

To bridge this gap, we propose \textbf{\textit{\NickName-Bench}}, a comprehensive benchmark specifically designed to evaluate the spatial intelligence capabilities of VLMs in UAV navigation scenes. 
As shown in Fig.~\ref{fig1} and Fig.~\ref{fig: distribution}, our benchmark covers two main categories and thirteen fine-grained sub-capabilities, systematically evaluating the VLM's understanding of UAV scenes. 
The first category is \textbf{\textit{environmental perception}} capabilities, including (1) bounding box localization for accurate object detection; (2) target color recognition from a UAV perspective; (3) distance estimation between objects; (4) height perception from UAV view; (5) pointing to objects to locate the coordinates of a specific target; (6) pointing in reverse to identify objects at a given coordinate location; (7) free space detection to identify navigable areas; and (8) spatial relationship understanding to determine the relative positions between targets. 
The second category is \textbf{\textit{scene understanding}}, assessing the advanced cognitive abilities necessary for autonomous UAV navigation, including (9) scene captioning of a single UAV-view image, (10) time-series captioning of multiple images, (11) functional reasoning of objects in an image, (12) object counting at different scales and under occlusion, and (13) integrating overall spatial cues to determine whether a location is suitable for UAV landing.

To enhance the UAV spatial intelligence capabilities of VLM, we propose a scalable data generation method and introduce the \textbf{\textit{\NickName-Dataset}}, a training dataset containing 1 M samples with diverse question-answer templates. 
Our data generation process utilizes multimodal inputs, including RGB images, semantic mask labels, LiDAR depth data, pose information, and bounding box annotations, to automatically generate origin dataset and question-answer pairs covering all 13 benchmark tasks. 
We then trained \textbf{\textit{Sky-VLM}}, a spatial specific VLM for UAV navgation, using a two-stage training approach: first, we use supervised fine-tuning (SFT) on the \textbf{\textit{\NickName-Dataset}} to acquire UAV-specific spatial reasoning capabilities; then, we add reinforcement fine-tuning (RFT) using Group Relative Policy Optimization (GRPO) ~\cite{grpo,zhang2025lips,team2025robobrain,wu2025evaluating} to further optimize the model's performance in key spatial reasoning tasks like box, pointing and counting. 
Extensive experiments demonstrate that \textbf{\textit{Sky-VLM}} achieves state-of-the-art (SOTA) performance across all \textbf{\textit{\NickName-Bench}} tasks, significantly outperforming existing open-source and closed-source VLMs in UAV scene spatial intelligence.

Our main contributions are summarized as follows:
\begin{itemize}
\item We propose \textbf{\textit{\NickName-Bench}}, a comprehensive benchmark covering 2 main categories and 13 fine-grained sub-capabilities, for systematically evaluating the spatial intelligence of VLMs in UAV navigation scenarios.

\item We construct \textbf{\textit{\NickName-Dataset}}, a large-scale dataset containing 1 M samples generated through an automated process, including various annotation formats including open question-answer, multiple choice, pointing, and bounding boxes, covering all benchmark tasks.

\item We propose Sky-VLM, a dedicated UAV-view spatial awareness VLM trained using a two-stage approach: first, SFT to acquire UAV-specific spatial reasoning capabilities, and then GRPO to enhance its decision-making ability in complex navigation scenarios.

\item Extensive experiments demonstrate that \textbf{\textit{Sky-VLM}} achieves SOTA performance across all \textbf{\textit{\NickName-Bench}} tasks, significantly outperforming both open-source and closed-source VLMs.

\end{itemize}

\section{Related Work}
\label{sec:related}

\textbf{VLM for UAV Navigation}
Unmanned aerial vehicle (UAV) navigation aims to enable UAVs to navigate autonomously based on high-level human commands.
Traditional methods rely on supervised learning using human commands and flight trajectories collected in specific scenarios~\cite{Giusti2016,sadeghi2017cad2rl}. Recently, VLM with its powerful visual language understanding capabilities has significantly advanced UAV navigation tasks~\cite{zhang2025mapnav,zhang2025nava,team2025robobrain}.
For example, UAV-VLA~\cite{uavvla} combines satellite imagery with the inference capabilities of VLMs to generate mission plans. See, Point, Fly, seepointfly~\cite{seepointfly} maps the output of VLMs to 3D waypoints through an intuitive visual pointing interface, achieving zero-shot UAV navigation. SoraNav~\cite{soranav} enriches the VLM input with geometric priors and switches between VLM inference and geometry-driven exploration based on navigation history; VLM-RRT~\cite{vlmrrt} utilizes directions proposed by VLMs to guide RRT* sampling, accelerating path convergence.
However, readily available VLMs generally lack accurate spatial awareness of UAV scenarios. For UAV navigation, spatial intelligence is crucial. It requires models to deeply understand spatial relationships, perform fine-grained scene analysis, and achieve accurate environmental perception, thereby supporting real-time flight decisions.

\textbf{Spatial Intelligence Benchmark}
% While early vision-language benchmarks like VQA~\cite{vqa} and GQA~\cite{gqa} focus on semantic reasoning from static, ground-level images, they lack the capacity to evaluate fine-grained visual-spatial intelligence---the ability to mentally reconstruct environments and reason about spatial relationships. Recent benchmarks address this gap: VSI-Bench~\cite{VSI} uses indoor videos to assess dynamic spatial memory over time, MMSI-Bench~\cite{mmsi} evaluates multi-image spatial reasoning across diverse domains with complex, multi-step questions, RynnEC-Bench~\cite{dang2025rynnec} offers 22 region-centric embodied tasks from 20K+ egocentric videos and RefSpatial-Bench~\cite{zhou2025roborefer} introduces multi-step spatial referring (up to 5 steps) with 200 real-scene masks.
% However, all of these---and indeed all existing spatial benchmarks---remain confined to terrestrial or egocentric viewpoints that mirror human visual experience, overlooking key challenges of aerial perspectives such as extreme scale variation, top-down occlusions, missing depth cues, and the need for metric-aware grounding. These limitations hinder reliable deployment in real-world UAV scenarios where precise spatial understanding from above is essential. To the best of our knowledge, no benchmark supports fine-grained evaluation of visual-spatial intelligence from an aerial viewpoint. We bridge this gap with \textbf{\textit{\NickName-Dataset}}, the first benchmark designed to assess multimodal large language models' spatial perception and scene-level reasoning in drone-captured environments.
Recently, several spatial intelligence benchmarks have emerged to evaluate the spatial perception capabilities of VLMs for a wide range of tasks~\cite{VSI,mmsi,dang2025rynnec,zhou2025roborefer,Robospatial,fu2024blinkmultimodallargelanguage,tong2024cambrian1fullyopenvisioncentric,du2024embspatialbenchbenchmarkingspatialunderstanding,geminiroboticsteam2025geminiroboticsbringingai,wang2024allseeingprojectv2general,wang2025embodiedsceneunderstandingvision,ray2024sat,yin2025spatial,yang2025thinking,zhu2025cvbench}. These benchmarks each have their own focus: early benchmarks like VQA~\cite{vqa} and GQA~\cite{gqa} emphasized semantic reasoning from static ground images. VSI-Bench~\cite{VSI} used indoor video to assess the changes in dynamic spatial memory over time. MMSI-Bench~\cite{mmsi} tested spatial reasoning across multiple images using complex multi-step problems. RynnEC-Bench~\cite{dang2025rynnec} extracted 22 region-based embodied cognition tasks from massive amounts of egocentric video. RefSpatial-Bench~\cite{zhou2025roborefer} introduced a spatial reference task involving up to five steps. RoboSpatial~\cite{Robospatial} provided a large-scale 2D/3D dataset with multi-perspective spatial annotations specifically for robotics-oriented spatial understanding,while Blink~\cite{fu2024blinkmultimodallargelanguage} highlighted that even basic perceptual capabilities underpinning such reasoning—like relative depth and visual correspondence—remain poorly supported by current models.
However, all existing spatial intelligence benchmarks share a common limitation: they all focus on spatial perception from a ground-based or egocentric perspective, neglecting the perception challenges from a UAV's perspective. For example, there are challenges such as varying object scales, top-down occlusion, lack of depth information, and complex ground understanding requirements. To address this gap, we propose \textbf{\textit{\NickName-Bench}} for evaluation and \textbf{\textit{\NickName-Dataset}} for training.

\section{\NickName Dataset and Benchmark Construction}
\label{method}
\begin{figure*}
\centering
    \captionsetup{type=figure}
    % \vspace{-32pt}
    \includegraphics[width=0.94\textwidth]{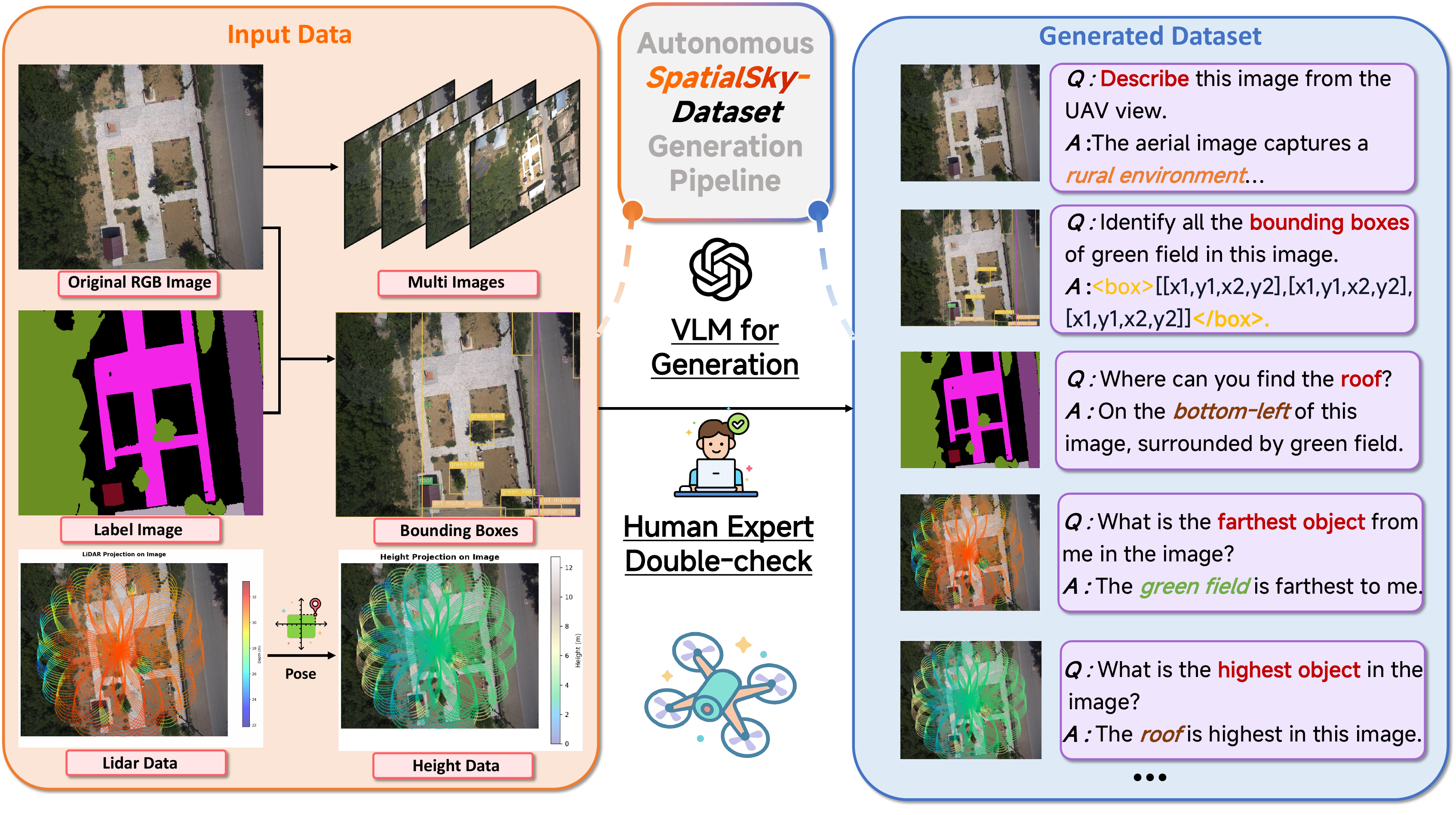}
    \vspace{-0.6em}
\captionof{figure}{\textbf{\textit{\NickName-Dataset} Generation Process.} 
Our generation pipeline take multimodal inputs, including RGB images, semantic labels, LiDAR depth data, UAV pose information, and bounding boxes. Using a VLM-based generation method and human expert validation, we automatically generate diverse question-answer pairs for 13 spatial reasoning tasks.}
\vspace{-1.1em}
    \label{fig2}
\end{figure*}

\subsection{Data Collection and Filtering}
% The \NickName-Dataset integrates annotations from
% UAVScenes[cite], which include 20k images and 对应的雷达数据和mask级别的图像中的类别标注，包含22个物体类别. By combining 2d image mask and 3d lidar data, we create a diverse dataset for modeling
% object and spatial interactions. This dataset consists of two
% main components: Environmental Perception and Scene Understanding.
The \textbf{\textit{\NickName-Dataset}} integrates annotations from UAVScenes~\cite{wang2025uavscenes}, which include 20,000 images along with corresponding radar data and class labels at the mask level, covering 22 object categories. By combining 2D image masks with 3D LiDAR data, we create a diverse dataset for modeling object and spatial interactions. This dataset consists of two main components: \textbf{\textit{Environmental Perception}} and \textbf{
\textit{Scene Understanding}}.

\textbf{Environmental Perception}
Our environment-aware dataset incorporates eight fine-grained spatial reasoning capabilities, leveraging multimodal inputs including RGB images, semantic segmentation masks, LiDAR point clouds, and UAV pose information.
For \textit{bounding box localization}, we directly extract object instances from pixel-level semantic masks and transform each connected component into an axis-aligned bounding box $(x_1, y_1, x_2, y_2)$. 
% To ensure class balance among the 22 object categories, we apply stratified sampling to maintain a balanced ratio of common to rare categories.
For \textit{color recognition}, we analyze the RGB distribution in each segmentation mask, calculating the dominant color by clustering pixel values in the HSV space and mapping them to descriptors like ``light blue." 
For pointing and reverse pointing tasks, we sample 5--8 pixel coordinates $(x_i, y_i)$ within each mask region. The \textit{reverse pointing} task uses these coordinates to identify the corresponding object category.
For \textit{free space detection and spatial relationship inference}, we utilize geometric properties of segmentation masks. We identify sufficiently large ($>$500 pixels) connected background regions and extract 3--5 points from each region.
% For \textit{color recognition}, we analyze the RGB distribution within each segmentation mask region, calculating the dominant color by clustering pixel values in the HSV color space and mapping them to natural language descriptors like ``light blue''. 
% For pointing and reverse pointing tasks, we sample multiple reference points within each mask region. Specifically, for each object instance, we randomly select 5–8 pixel coordinates $(x_i, y_i)$ within the mask boundaries. 
% The \textit{reverse pointing} task uses these same coordinates to generate the corresponding object category.
% For \textit{free space detection and spatial relationship inference}, we utilize the geometric properties of segmentation masks. The free space identification method involves detecting sufficiently large ($>$500 pixels) connected background regions and extracting 3-5 points within each region. 
%%%%%%%%%%%%%%%%%%%%%%%%
Spatial relationship inference is achieved by calculating the relative positions of object pairs. Given masks $M_i$ and $M_j$, we compute their centroids $c_i = (\bar{x}_i, \bar{y}_i)$ and $c_j = (\bar{x}_j, \bar{y}_j)$, then determine the directional relationship using angle and distance thresholds:
% Spatial relationship inference is achieved by calculating the relative positions between object pairs. Given two masks $M_i$ and $M_j$, we compute their centroids $c_i = (\bar{x}_i, \bar{y}_i)$ and $c_j = (\bar{x}_j, \bar{y}_j)$, and then determine the directional relationship based on angle and distance thresholds:
\vspace{-0.5em}
\begin{equation}
    \theta_{ij} = \arctan\left(\frac{\bar{y}_j - \bar{y}_i}{\bar{x}_j - \bar{x}_i}\right), \quad d_{ij} = \|c_i - c_j\|_2.
    \vspace{-0.5em}
\end{equation}
When $d_{ij}$ exceeds a minimum threshold of 50 pixels, we categorize the relations into eight classes (left, right, top, etc.) and generate our original spatial relation dataset.
For \textit{distance estimation}, we directly utilize the LiDAR point cloud: $\mathbf{P} = \{p_k\}_{k=1}^N$, where each point $p_k = (x_k, y_k, z_k)$ lies in the LiDAR coordinate system.
We project the LiDAR points onto the image plane using the camera intrinsic parameters $\mathbf{K}$ and extrinsic parameters $[\mathbf{R}|\mathbf{t}]$, and then calculate the average depth:
\begin{equation}
    d_{\text{obj}} = \frac{1}{|\mathcal{P}_{\text{obj}}|} \sum_{p_k \in \mathcal{P}_{\text{obj}}} z_k^{\text{cam}},
\end{equation}
where $z_k^{\text{cam}}$ represents the depth value after converting the LiDAR points to camera coordinates. $p_k^{\text{cam}} = \mathbf{R}$.
To perform \textit{height estimation}, we use the UAV pose transformation moments $\mathbf{T}_{4\times4}$ to convert the LiDAR point array into world coordinates. The global altitude of each point is obtained using the following formula:
\vspace{-0.5em}
\begin{equation}
    \begin{bmatrix} x_k^w \\ y_k^w \\ z_k^w \\ 1 \end{bmatrix} = \mathbf{T}_{4\times4} \begin{bmatrix} x_k^{\text{cam}} \\ y_k^{\text{cam}} \\ z_k^{\text{cam}} \\ 1 \end{bmatrix},
    \vspace{-0.5em}
\end{equation}
where $z_k^w$ represents the absolute altitude in the world coordinate system. 
% For each object, we calculate the average height $h_{\text{mean}} = \frac{1}{|\mathcal{P}_{\text{obj}}|}\sum_{p_k \in \mathcal{P}_{\text{obj}}} z_k^w$ and the maximum height $h_{\text{max}} = \max_{p_k \in \mathcal{P}_{\text{obj}}} z_k^w$ to generate a height comparison query. We use a 6×4 grid overlaid on the image to cluster points into spatial regions, retaining only regions containing at least 20 valid LiDAR measurements to ensure reliable height estimation. 
For each object, we calculate the average height to generate a height comparison query. 
% This process generates question-and-answer pairs to identify the highest/lowest object and compares the relative height between object pairs when $|h_{\text{mean}}^i - h_{\text{mean}}^j| > 2.0$ meters.

\textbf{Scene Understanding}
Our scene understanding dataset comprises five high-level cognitive tasks that require holistic reasoning about aerial scenes, fully leveraging visual content and semantic context.
For \textit{single-image and multi-image captions}, we extract real-world object categories from semantic masks and input single-image and multi-image sequences into a VLM, providing cues that emphasize aerial perspective features. The model generates captions covering scene composition, spatial object distribution, environmental context, and spatial variations within the multi-image sequence.
For \textit{bject counting}, we apply connected component analysis to the semantic masks to identify individual instances $\{M_c^1, M_c^2, ..., M_c^{n_c}\}$ for each category $c$. To ensure class balance, we employ stratified sampling, oversampling rare categories and undersampling major categories.
For \textit{function reasoning}, we manually craft 2–3 functional descriptions for each of the 22 object categories to reflect realistic drone scenes. We combine backpointing with functional queries by sampling points $(x, y)$ within the object mask and generating questions. For landing safety analysis, we extract target distribution, available airspace ($>$1,000 pixels), potential hazards, and surface features, and then input them into the VLM. The model outputs a structured assessment, including feasibility classification (safe/cautious/unsafe), confidence score, recommended landing area, identified hazards and their risk levels, and comprehensive safety inference.

\begin{figure*}[!t]
\centering
    \captionsetup{type=figure}
    % \vspace{-32pt}
    \includegraphics[width=0.95\textwidth]{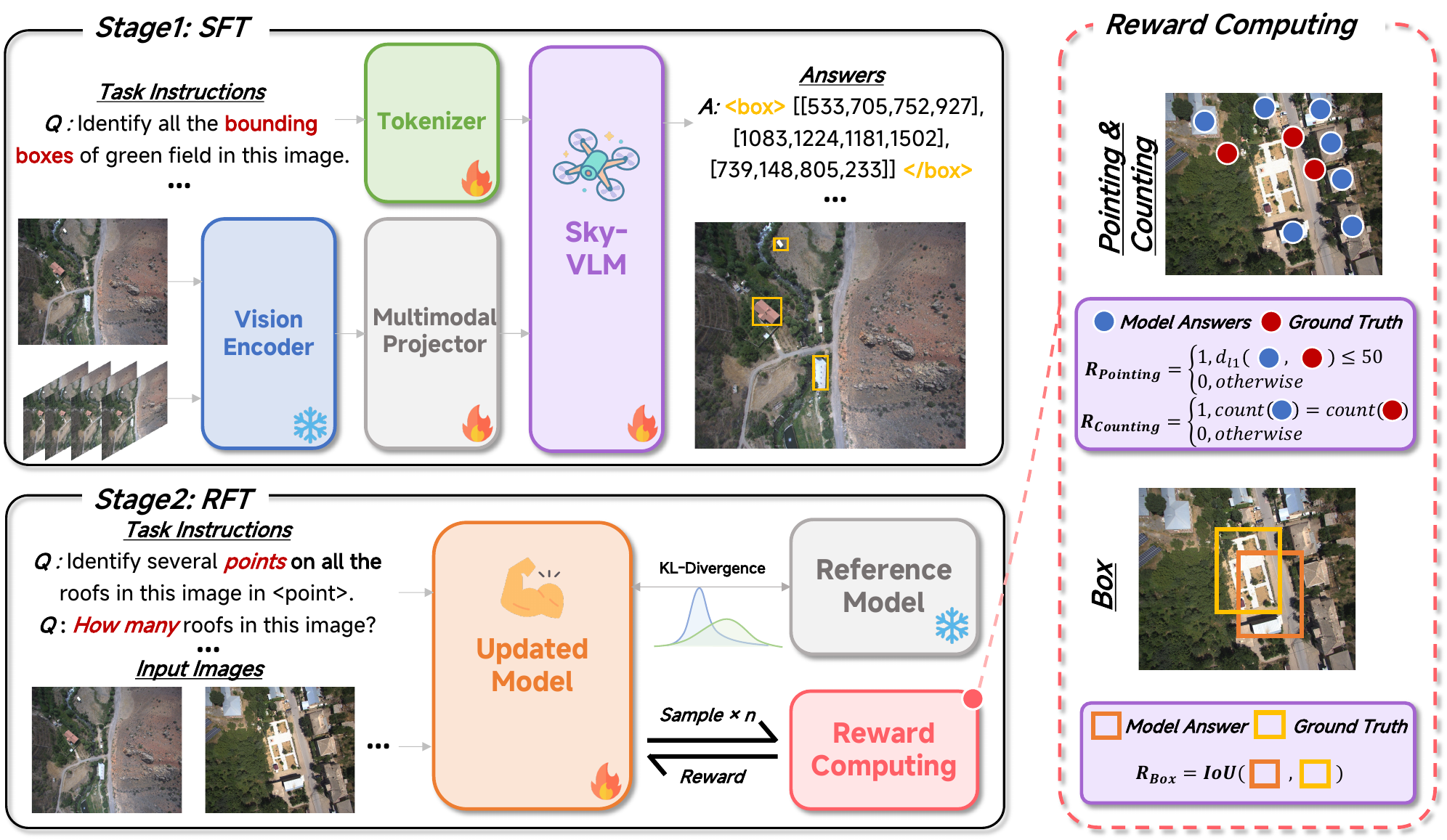}
    \vspace{-0.8em}
\captionof{figure}{\textbf{Overview of our Sky-VLM.} Sky-VLM adopts a two-stage training approach. In the first stage, we involve supervised fine-tuning (SFT) on the entire \NickName-Dataset to develop the basic spatial reasoning capabilities. In the second stage, we use reinforcement fine-tuning (RFT), utilizing task-specific reward functions to enhance decision-making accuracy for key spatial tasks.}
\vspace{-1em}
    \label{fig3}
\end{figure*}

\subsection{Question-Answer Pairs Generation}
To ensure diversity and robustness of the benchmark, we designed task-specific QA formats for 13 fine-grained spatial reasoning tasks. We use VLM to generate more than 20 different question templates for each task, covering a variety of linguistic expressions and query structures to prevent the model from relying on a single pattern match. 
% For example, bounding box localization queries include variations such as ``Identify all bounding boxes of [object]'', ``Locate [object] in this image”, and “Where is [object] located?”.

For the answer format, we adopted a structured representation to facilitate automatic evaluation. For bounding box and pointing tasks, we encapsulated coordinates in special tags: \texttt{<box>[[$x_1,y_1,x_2,y_2$],[$x_1,y_1,x_2,y_2$]]</box>} and \texttt{<point>[[$x_1,y_1$],[$x_2,y_2$]]</point>}. For color recognition, spatial relationships, and counting tasks, we constructed multiple-choice questions with 4–6 carefully crafted distractors and placed the answers within the \texttt{<//boxed><choice>} tag. For distance estimation, height comparison, reverse pointing, scene caption, function, and landing safety tasks, we use open-ended questions with no strict restrictions, allowing for free responses. This multi-question design ensures that our benchmark comprehensively assesses structured spatial reasoning and open-ended semantic understanding capabilities.

\subsection{\NickName-Bench}
To construct a comprehensive and impartial evaluation benchmark, we carefully selected approximately 1,000 QA pairs from the generated dataset, ensuring balanced coverage of all 13 fine-grained tasks, 22 object categories, and various scene types. We employed stratified sampling to guarantee representativeness for each task category and scene context. Crucially, after selecting these benchmark samples, we removed all other QA pairs associated with the same images from the training dataset, ensuring the benchmark consists entirely of unseen images to prevent data leakage and guarantee the fairness of the evaluation.

We designed task-specific evaluation metrics for each spatial reasoning ability. For bounding box localization, we calculate the Intersection over Union (IoU) between the predicted bounding box $B_{\text{pred}}$ and the ground truth bounding box $B_{\text{gt}}$, and average it over all instances:
\vspace{-0.5em}
\begin{equation}
    \text{mIoU} = \frac{1}{N}\sum_{i=1}^N \frac{|B_{\text{pred}}^i \cap B_{\text{gt}}^i|}{|B_{\text{pred}}^i \cup B_{\text{gt}}^i|}.
    \vspace{-0.5em}
\end{equation}
If $\text{IoU} \geq 0.5$, the prediction is considered correct. For pointed tasks, we evaluate whether the predicted point falls within the true target mask $M_{\text{gt}}$.
For multiple-choice questions and target object category recognition, we calculate the standard accuracy. For open-ended tasks, we use BLEU~\cite{papineni2002bleu} and GPT-4o~\cite{gpt4o} as an automatic evaluator, providing it with the question, the true answer, and the model prediction, and then asking it to give a score from 1 to 10 based on factual correctness, semantic completeness, and reasoning quality. The final score is the average of all samples: $\text{Score}_{\text{open}} = \frac{1}{N}\sum_{i=1}^N s_i$, where $s_i \in [1, 10]$.

\section{Sky-VLM}
\textbf{Framework}
% We propose Sky-VLM, a VLM specifically designed for UAV spatial reasoning tasks. As shown in Figure 3, our model employs a multimodal architecture, including a vision encoder, a multilayer perceptron (MLP) projector, a text segmenter, and the Qwen2.5-7B ~\cite{qwen} language model as its backbone. The vision encoder is designed to flexibly handle single-image and multi-image inputs: for single-image tasks, we process one frame of image through the encoder; for time-series tasks, we independently encode up to 10 consecutive frames and concatenate their features along the sequence dimension. The vision encoder extracts visual features from the input image and then transforms them into the same embedding space as the language tokens through the MLP projector. These visual embeddings are concatenated with embedded text instructions and fed into the LLM for joint reasoning across vision and language modalities, enabling the model to generate structured outputs or natural language responses according to task requirements.
We propose ~\textbf{\textit{Sky-VLM}}, a VLM built on Qwen2.5-VL-7B~\cite{bai2025qwen2} specifically designed for UAV spatial reasoning tasks. The model employs a multimodal architecture flexibly handling both single-image and multi-image inputs for UAV spatial reasoning.

\textbf{Supervised Fine-Tuning}
In the first phase of training, we performed supervised fine-tuning (SFT) on the entire \textbf{\textit{\NickName-Dataset}} containing 1 million samples to establish the foundational spatial reasoning capabilities for UAVs. This phase enabled the model to: (1) learn aerial visual representations distinct from ground-based perspectives; (2) acquire task-specific output formats, including structured coordinates (<box>, <point>), multiple choice (<boxed>), and free descriptions; (3) develop basic spatial reasoning capabilities across all 13 benchmark tasks.
We employed standard language modeling next-word prediction loss, but only computed gradients for the answer word to focus the learning on generating responses rather than understanding the question. Given a visual embedding sequence $\mathbf{V} = \{v_1, ..., v_m\}$ and a text tag sequence $\mathbf{T} = \{t_1, ..., t_n\}$, where the answer starts at position $k$, the SFT loss function can be expressed as:
\vspace{-0.5em}
\begin{equation}
    \mathcal{L}_{\text{SFT}} = -\frac{1}{n-k+1}\sum_{i=k}^{n} \log P(t_i | \mathbf{V}, t_1, ..., t_{i-1}; \theta),
    \vspace{-0.5em}
\end{equation}
where $\theta$ represents the model parameters, and $P(t_i | \cdot)$ represents the probability of predicting tag $t_i$ given the visual context and preceding tags.

\begin{table*}[!ht]
\centering

\fontsize{13}{16}\selectfont 
\resizebox{\textwidth}{!}{
\begin{tabular}{llcccccccccccccc}  % 16列
\toprule
\multirow{2}{*}{\textbf{Model}} & \multirow{2}{*}{\textbf{Params}} & \multicolumn{8}{c}{\textbf{Environmental Perception}} & \multicolumn{5}{c}{\textbf{Scene Understanding}} & \multirow{2}{*}{\textbf{Avg.}$\uparrow$} \\
\cmidrule(lr){3-10} \cmidrule(lr){11-15}
& & \cellcolor{gray!10}Box & \cellcolor{gray!10}Color & \cellcolor{gray!10}Dist. & \cellcolor{gray!10}Height & \cellcolor{gray!10}Point & \cellcolor{gray!10}Rev. & \cellcolor{gray!10}Free. & \cellcolor{gray!10}Sp. Rel. & \cellcolor{gray!10}Single & \cellcolor{gray!10}Multi & \cellcolor{gray!10}Cou. & \cellcolor{gray!10}Fun. & \cellcolor{gray!10}Land. & \\
\midrule
\rowcolor{red!10}\multicolumn{16}{c}{\textbf{\textit{Closed-source Models}}} \\
\midrule
GPT-4-mini~\cite{gpt4o} & - & 0.91 & 54.00 & 35.00 & 27.00 & 5.62 & 11.00 & 2.46 & 15.00 & 14.04 & 13.28 & 28.00 & 20.88 & 35.20 & 20.11 \\
GPT-4o~\cite{gpt4o} & - & 0.24 & 45.00 & 26.00 & 23.00 & 4.83 & 9.00 & 9.61 & 16.00 & 17.41 & 14.75 & 32.00 & 28.22 & 50.70 & 21.27 \\
GPT-5~\cite{gpt4o} & - & 1.13 & 47.00 & 35.00 & 33.00 & 1.38 & 11.00 & 5.03 & 27.00 & 10.51 & 10.62 & 27.00 & 40.81 & 50.50 & 23.07 \\
Gemini-2.5-Flash~\cite{comanici2025gemini} & - & 2.10 & 38.00 & 25.00 & 48.00 & 5.05 & 12.00 & 14.71 & 11.00 & 11.77 & 8.78 & 37.00 & 24.62 & 47.90 & 21.99 \\
Gemini-2.5-Pro~\cite{comanici2025gemini} & - & 3.45 & 46.00 & 24.00 & 40.00 & 6.45 & 9.00 & 12.39 & 24.00 & 12.21 & 11.24 & 24.00 & 37.12 & 46.30 & 22.75 \\
Qwen-VL-Max~\cite{bai2025qwen2} & - & 1.50 & 48.00 & 34.00 & 29.00 & 1.21 & 8.00 & 0.00 & 24.00 & 13.47 & 14.08 & 29.00 & 15.58 & 52.40 & 20.77 \\
\midrule
\rowcolor{yellow!10}\multicolumn{16}{c}{\textbf{\textit{Open-source Models}}} \\
\midrule
InternVL3.5~\cite{wang2025internvl3} & 8B & 1.41 & 47.00 & 26.00 & 24.00 & 4.48 & 7.00 & 8.64 & 32.00 & 13.64 & 11.19 & 12.00 & 20.05 & 35.40 & 18.65 \\
Qwen3-VL-8B~\cite{yang2025qwen3} & 8B & 1.32 & 12.00 & 26.00 & 44.00 & 6.46 & 3.00 & 6.93 & 33.00 & 11.07 & 12.30 & 12.00 & 26.16 & 4.33 & 15.25\\
Qwen2.5-VL-7B~\cite{bai2025qwen2} & 7B & 2.38 & 46.00 & 17.00 & 25.00 & 0.79 & 9.00 & 0.00 & 27.00 & 15.27 & 12.33 & 18.00 & 17.43 & 31.32 & 16.93 \\
Qwen2.5-VL-32B~\cite{bai2025qwen2} & 32B & 0.27 & 47.00 & 16.00 & 13.00 & 9.88 & 6.00 & 2.34 & 21.00 & 10.78 & 6.98 & 0.00 & 32.67 & 15.50 & 13.93 \\
Qwen2.5-VL-72B~\cite{bai2025qwen2} & 72B & 0.23 & 53.00 & 23.00 & 25.00 & 11.08 & 6.00 & 3.25 & 22.00 & 12.22 & 10.08 & 0.00 & 33.08 & 10.11 & 16.05 \\
\midrule
\rowcolor{blue!10}\multicolumn{16}{c}{\textbf{\textit{Spatial Specific Models}}} \\
\midrule
SpatialVLM~\cite{chen2024spatialvlm} & 8B & 0.96 & 21.00 & 52.00 & 58.00 & 0.85 & 11.00 & 0.00 & 25.00 & 12.05 & 10.30 & 13.00 & 19.08 & 25.90 & 19.02 \\
SpaceR~\cite{spacer} & 7B & 7.45 & 4.00 & 27.00 & 35.00 & 0.64 & 3.00 & 4.55 & 29.00 & 13.93 & 10.63 & 4.00 & 22.45 & 2.96 & 12.61 \\
VILASR~\cite{vilasr} & 7B & 2.08 & 0.00 & 37.00 & 42.00 & 3.92 & 7.00 & 3.87 & 26.00 & 12.92 & 12.92 & 0.00 & 24.33 & 2.76 & 13.45 \\
\rowcolor{myblue}\textbf{Sky-VLM (Ours)} & 7B & \textbf{42.68} & \textbf{79.00} & \textbf{84.00} & \textbf{79.00} & \textbf{30.72} & \textbf{60.00} & \textbf{43.20} & \textbf{64.00} & \textbf{27.34} & \textbf{23.83} & \textbf{52.00} & \textbf{45.72} & \textbf{61.40} & \textbf{53.30} \\
\bottomrule
\end{tabular}
}
\vspace{-0.5em}
\caption{\textbf{Comparison Results of Various VLMs on \NickName-Bench.} Our \textbf{\textit{Sky-VLM}}  achieves SOTA performance. Dist., Rev., Free., Sp. Rel., Cou., Fun., Land., Avg., denote distance, reverse point, freespace, spatial relation, counting, function, landing and total average.}
\label{tab:model_performance}
\vspace{-1em}
\end{table*}
% \vspace{-1em}

\textbf{Reinforcement Fine-Tuning}
In the second stage, we apply Group Relative Policy Optimization (GRPO) to further improve the model's decision-making ability and output accuracy in key spatial reasoning tasks. We constructed a reinforcement learning dataset containing 30,000 samples, focusing on tasks requiring precise localization and structured output. As shown in Fig.~\ref{fig3}, we designed a task-specific reward function to directly measure the deviation between the model's predictions and the ground truth labels.

For the pointing task, we evaluate the sequence of points output by the model, and the final task score is the average of the scores of all predicted points. The reward for a single point is binary, determined by calculating the L1 distance between the predicted point $(x_{\text{pred}}, y_{\text{pred}})$ and its nearest ground truth point $(x, y)$, using the following criteria:
\vspace{-0.5em}
\begin{equation}
    R_{\text{point}} = \begin{cases} 1, & \text{if }  |x_{\text{pred}} - x| + |y_{\text{pred}} - y| \leq 50, \\ 0, & \text{otherwise}. \end{cases}
    \vspace{-0.2em}
\end{equation}
For multiple-choice tasks, we use an exact match reward:
\begin{equation}
   R_{\text{choice}} = \begin{cases} 1, & \text{if predicted answer} = \text{true answer}, \\ 0, & \text{otherwise}.\end{cases} 
\end{equation}
For bounding box localization, we calculate the IoU between the predicted and ground truth boxes as a continuous reward signal:
\vspace{-0.5em}
\begin{equation}
  R_{\text{box}} = \text{IoU}(B_{\text{pred}}, B_{\text{gt}}) = \frac{|B_{\text{pred}} \cap B_{\text{gt}}|}{|B_{\text{pred}} \cup B_{\text{gt}}|}. 
  \vspace{-0.5em}
\end{equation}
The GRPO objective function aims to maximize the expected reward while maintaining closeness to the reference model $\pi_{\text{ref}}$ through KL divergence regularization:
\vspace{-0.5em}
\begin{equation}
    \mathcal{L}_{\text{GRPO}} = -\mathbb{E}_{\pi_{\theta}}\left[R(y) \cdot \log \frac{\pi_{\theta}(y|x)}{\pi_{\text{ref}}(y|x)}\right] + \beta \cdot \text{KL}(\pi_{\theta} || \pi_{\text{ref}}),
    \vspace{-0.3em}
\end{equation}
where $\beta$ controls the strength of the KL penalty. Through the reinforcement learning phase, \textbf{\textit{Sky-VLM}} is able to learn to generate more accurate spatial localization and consistent structured output, especially in tasks requiring pixel-level accuracy, where performance is significantly improved.

\section{Experiments}

\subsection{Experimental Setup}
\textbf{Implementation Details}
Our \textbf{\textit{Sky-VLM}} model is based on the Qwen2.5-VL-7B~\cite{bai2025qwen2} and employs a two-stage process for initialization and training. The first stage involves supervised SFT on the \textbf{\textit{\NickName-Dataset}} containing 1M samples. This stage utilizes eight H200 GPUs, the AdamW optimizer~\cite{loshchilov2017decoupled}, a learning rate of 1e-5, a batch size of 2 per device, 2 gradient accumulation steps, and trains for one epoch. The second stage employs RFT using the GRPO~\cite{grpo} algorithm, trained on a dedicated dataset of 30K samples. In the RFT stage, a model with a learning rate of 1e-6 and weight decay of 0.1 is trained for one epoch, using the SFT model as the reference policy, with a KL regularization coefficient ($\beta$) of 0.01 to improve decision accuracy.

\begin{figure}[!t]
\centering
    \captionsetup{type=figure}
    
    \includegraphics[width=0.48\textwidth]{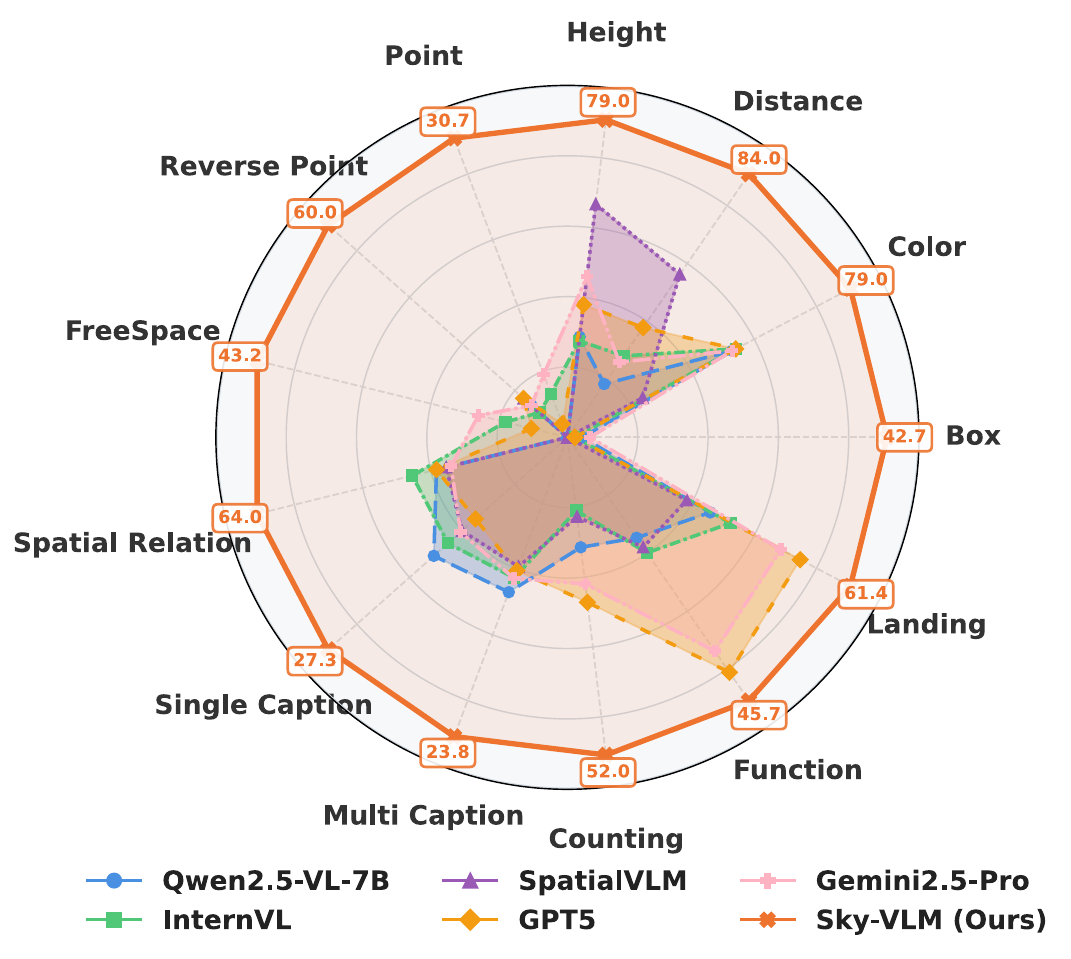}
    \vspace{-1.5em}
\captionof{figure}{\textbf{Performance of Our Sky-VLM.} }
\vspace{-1.5em}
    \label{fig: lidar}
\end{figure}

\subsection{Evaluation Metrics}
Our evaluation on \textbf{\textit{\NickName-Bench}} employs a range of task-specific metrics for different spatial reasoning capabilities. For bounding boxes, we compute mIoU. For pointing tasks, accuracy depends on whether the predicted coordinates lie within the mask of the real object. For tasks with discrete answers, such as multiple-choice questions and object category recognition, we report standard accuracy. Finally, for tasks such as image captioning and functional reasoning, we use BLEU-1 to BLEU-4 scores~\cite{papineni2002bleu}. For landing, we use GPT-4o~\cite{gpt4o} as an automated evaluator. The final score for these tasks is the average of all sample scores.

\subsection{Comparison with State-of-the-Art Models}
We compared \textbf{\textit{Sky-VLM}} with a comprehensive range of baseline models, including state-of-the-art closed-source models (GPT5~\cite{gpt4o}, Gemini2.5-Pro~\cite{comanici2025gemini}, \textit{etc.}), open-source general-purpose VLMs (InternVL3.5~\cite{wang2025internvl3}, Qwen2.5-VL~\cite{bai2025qwen2}, \textit{etc.}), and spatial specific models (SpatialVLM~\cite{chen2024spatialvlm}, etc.). As shown in Tab.~\ref{tab:model_performance} and Fig~\ref{fig: lidar}, existing models perform poorly across all spatial inference tasks. The average scores of closed-source models range from 20.11 to 23.07, while open-source VLMs perform even worse (13.93 to 18.65). Even spatial specific models fail to effectively transfer to UAV perspectives; for example, SpatialVLM~\cite{chen2024spatialvlm}, SpaceR~\cite{spacer}, and VILASR~\cite{vilasr} achieve scores of only 19.02, 13.59, and 13.45, respectively. In comparison, our \textbf{\textit{Sky-VLM}} model achieved SOTA performance across all the models, with an average score of \textbf{53.30}, \textbf{139.6\%} improvement over the best baseline model (GPT-5, 23.07). 
\textbf{\textit{Sky-VLM}} demonstrated superior performance across key tasks: bounding box score of 42.68 mIoU (\textbf{473\%} improvement over SpaceR~\cite{spacer}), color score of 79.00 (\textbf{58\%} improvement over SpatialVLM~\cite{chen2024spatialvlm}), spatial relationship score of 70.00 (\textbf{38\%} improvement over InternVL3.5~\cite{wang2025internvl3}), and landing score of 61.40 (\textbf{9\%} improvement over Qwen-VL-Max~\cite{bai2025qwen2}). This demonstrates the powerful spatial intelligence capabilities of our model in UAV scenarios.

\begin{table}[!t]
\centering

\fontsize{15}{20}\selectfont 
\resizebox{0.48\textwidth}{!}{
\begin{tabular}{lc|c|c|c}
\toprule
\multirow{2}{*}{\textbf{Model}} & \textbf{Params} & \textbf{Env. Per.} & \textbf{Sce. Und.} & \textbf{Total} \\
& &\textbf{Avg.}$\uparrow$ &\textbf{Avg.}$\uparrow$&\textbf{Avg.}$\uparrow$ \\
\midrule
\rowcolor{mygreen}\multicolumn{5}{c}{\textbf{\textit{Other Models}}} \\
\midrule
Gemini-2.5-Pro~\cite{comanici2025gemini} & - & 20.61 & 26.17 & 22.75 \\
Qwen2.5-VL-7B~\cite{bai2025qwen2} & 7B & 15.75 & 18.81 & 16.93 \\
SpaceR~\cite{spacer} & 7B & 13.75 & 10.84 & 12.78 \\
\midrule
\rowcolor{myred}\multicolumn{5}{c}{\textbf{\textit{Our Models}}} \\
\midrule
\textbf{Sky-VLM-SFT (Ours)} & 7B & 52.53 & 41.52 & 48.29 \\
\textbf{Sky-VLM-RL (Ours)} & 7B & \textbf{60.33} & \textbf{42.06} & \textbf{53.30} \\
\bottomrule
\end{tabular}

}
\vspace{-0.2em}
\caption{\textbf{Ablation Study of Multi-Stage Training.}}

\label{tab:ablation_stage}
\vspace{-1em}
\end{table}

\subsection{Ablation Study}
\textbf{Effect of Multi-Stage Training}
To verify the effectiveness of our proposed two-stage training method, we compared Sky-VLM-SFT, trained with only SFT, with Sky-VLM-RL, which incorporates reinforcement learning. As shown in Tab.~\ref{tab:ablation_stage}, adding GRPO-based RFT significantly improved performance on spatial reasoning tasks (\textbf{53.30} vs. 48.29). Sky-VLM-RL achieved 60.33 score on the environment perception task, a 14.8\% improvement over Sky-VLM-SFT (52.53), while maintaining similar performance on the scene understanding task (42.06 vs. 41.52).
% This demonstrates that the reinforcement fine-tuning stage using a task-specific reward function effectively improves the model's accuracy on structured spatial reasoning tasks (e.g., bounding box localization, pointing, and counting) without sacrificing its general scene understanding capabilities. The overall performance improved from 48.29% to 53.30% (a relative improvement of 10.4%), validating the effectiveness of our proposed two-stage training strategy for UAV spatial intelligence.

\begin{figure*}[!t]
\centering
    \captionsetup{type=figure}
    
    \includegraphics[width=0.92\textwidth]{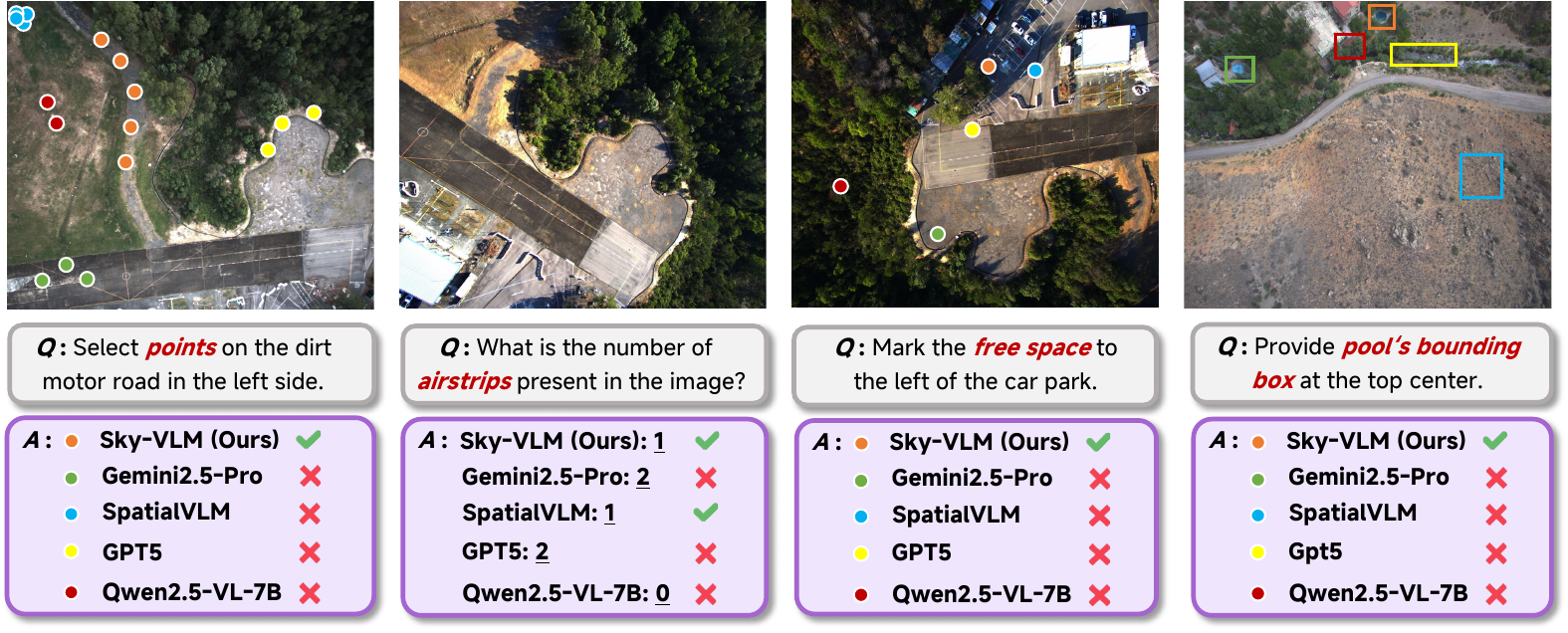}
    \vspace{-0.8em}
\captionof{figure}{\textbf{Qualitative Results of Different VLMs on \textit{\NickName-Bench}.} }

    \label{fig: qualitative}
    \vspace{-1.5em}
\end{figure*}

\textbf{Effect of Reward Model}
To demonstrate the role of each reward function in GRPO training, we conducted ablation experiments by removing individual reward components. As shown in Tab.~\ref{tab:ablation_reward}, all three reward functions play a crucial role in achieving optimal performance. Removing the point reward resulted in the most significant performance degradation, with the environment perception score dropping from 60.33 to 53.77 (6.56\% improvement), indicating that accurate coordinate prediction is fundamental to spatial reasoning. The bounding box reward and multiple-choice reward are equally important; removing them reduced the total average to 49.72 and 50.27, respectively, demonstrating their criticality for accurate object localization and discrete decision-making tasks such as color recognition and spatial relationships. 
% The complementarity of these reward functions—for coordinate accuracy, bounding box accuracy, and category reasoning, respectively—enables Sky-VLM-RL to achieve comprehensive spatial intelligence across all UAV navigation tasks.

\begin{table}[!t]
\centering

\fontsize{15}{20}\selectfont 
\resizebox{0.48\textwidth}{!}{
\begin{tabular}{l|c|c|c}
\toprule
\multirow{2}{*}{\textbf{Model}}  & \textbf{Env. Per.} & \textbf{Sce. Und.} & \textbf{Total} \\
&\textbf{Avg.}$\uparrow$&\textbf{Avg.}$\uparrow$&\textbf{Avg.}$\uparrow$ \\
\midrule
Qwen2.5-VL-7B~\cite{bai2025qwen2}  & 15.75 & 18.81 & 16.93 \\

Sky-VLM-SFT  & 52.53 & 41.52 & 48.29 \\
Sky-VLM-RL w/o Box Reward  & 57.66 & 41.78 & 49.72 \\
Sky-VLM-RL w/o Point Reward& 53.77 & 40.95 & 47.36 \\
Sky-VLM-RL w/o Multi-Choice Reward & 59.32 & 41.22 & 50.27 \\
\rowcolor{myblue}\textbf{Sky-VLM-RL (Ours)}  & \textbf{60.33} & \textbf{42.06} & \textbf{53.30} \\
\bottomrule
\end{tabular}
}
\vspace{-0.5em}
\caption{\textbf{Ablation Study of Reward Model.}}
\label{tab:ablation_reward}
\vspace{-0.8em}
\end{table}

\begin{figure}[t]
\centering
    \captionsetup{type=figure}
    
    \includegraphics[width=0.48\textwidth]{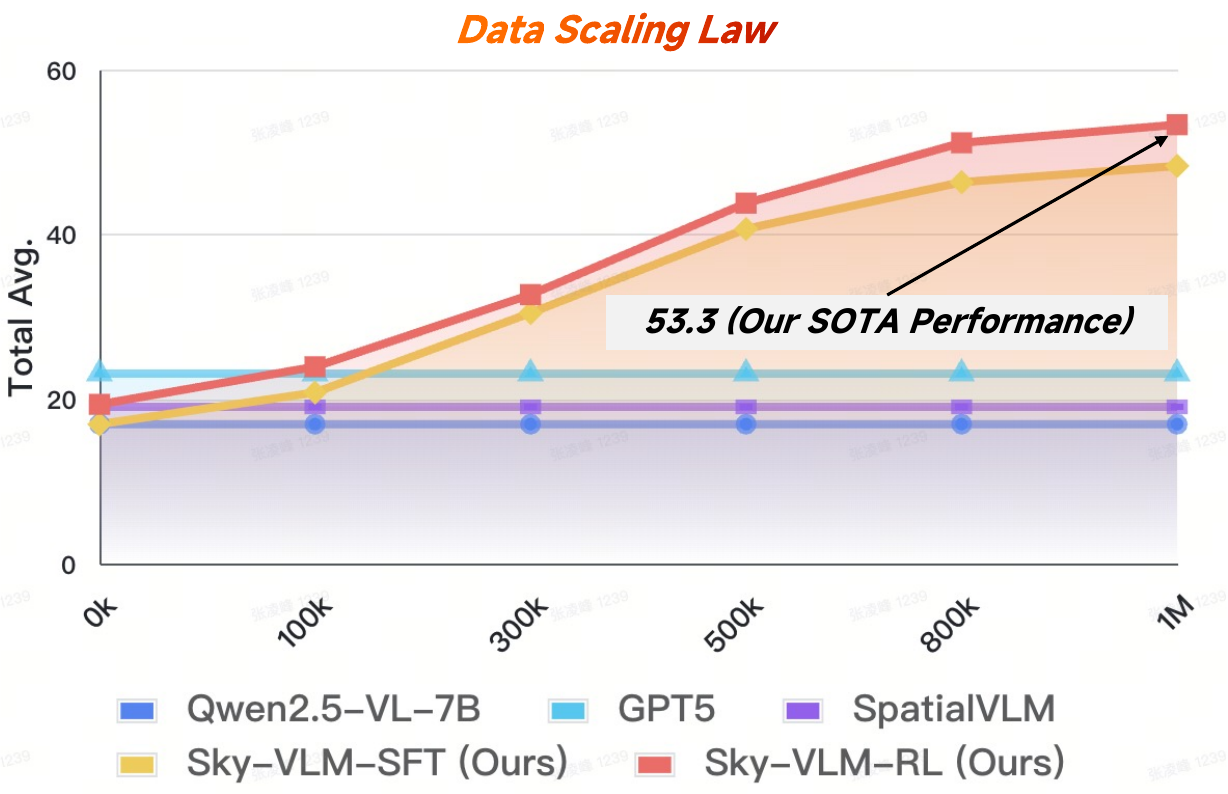}
    \vspace{-2em}
\captionof{figure}{\textbf{Data Scaling Law.} }
\vspace{-1em}
    \label{fig: ablation_data}
    \vspace{-1em}
\end{figure}

\textbf{Data Scaling Law}
To investigate the impact of training data size on model performance, we compared datasets with different sample sizes from 0K to 1M. As shown in Fig.~\ref{fig: ablation_data}, our model performance improved rapidly in the initial stage, then gradually saturated. Sky-VLM-SFT showed a significant improvement in the early stages, with accuracy jumping from 16.93 (baseline) to 30.43 using only 300K samples. Accuracy gradually decreased with increasing sample size, reaching 48.29 with 1 million samples. More importantly, the reinforcement learning stage consistently improved SFT performance across all data sizes. With 100K samples, Sky-VLM-RL achieved a score of 23.9, while SFT achieved 20.77 (3.13\% improvement); on the complete 1 million sample dataset, accuracy reached \textbf{53.3}, while SFT achieved 48.29 (\textbf{5.01\%} improvement). This continuous improvement gap demonstrates that RFT can effectively enhance spatial reasoning capabilities.

\subsection{Qualitative Analysis}

Fig.~\ref{fig: qualitative} presents a qualitative comparison of four representative spatial reasoning tasks, demonstrating that \textbf{\textit{Sky-VLM}} outperforms baseline models. In pointing task, Sky-VLM accurately identifies multiple valid locations on the road surface, while other models fail to provide correct predictions. In counting task, Sky-VLM correctly identifies one runway, while other models provide inaccurate counts, highlighting the challenge of object recognition from an aerial perspective. In freespace task, Sky-VLM successfully locates an open area, while all baseline models fail to complete this task. Furthermore, in box task, Sky-VLM generates an accurate bounding box at the top center, while baseline models are completely unable to detect or locate the target object.
These visualizations clearly demonstrate that Sky-VLM's spatial intelligence capabilities in UAV scenarios are significantly superior to other models.

\vspace{-2pt}
\section{Conclusion}
We propose \textbf{\textit{\NickName-Bench}}, which covers 13 fine-grained spatial reasoning tasks, categorized into environmental perception and scene understanding. Our extensive evaluation of mainstream VLMs reveals their significant limitations in spatial intelligence when handling UAV perspectives, highlighting the unique challenges posed by UAV navigation scenarios.
To address these challenges, we developed the \textbf{\textit{\NickName-Dataset}} dataset, containing 1 million automatically generated samples with diverse annotation methods, and propose \textbf{\textit{Sky-VLM}}. \textbf{\textit{Sky-VLM}} is a specific VLM trained using a two-stage approach, combining SFT and RFT, supplemented by task-specific rewards. Extensive experimental results demonstrate that \textbf{\textit{Sky-VLM}} achieves state-of-the-art performance across all benchmark tasks, significantly outperforming both open-source and closed-source VLMs in UAV spatial reasoning, paving the way for developing spatial-aware VLMs in UAV scenarios.
% \clearpage

{
    \small
    \bibliographystyle{ieeenat_fullname}
    
    \bibliography{main}
}


\end{document}